\documentclass[letterpaper]{article} 
\usepackage[preprint]{aaai2027}  
\usepackage[hyphens]{url}  
\usepackage{graphicx} 
\usepackage{natbib}  
\usepackage{caption} 
\usepackage{amsmath}
\usepackage{amssymb}
\usepackage{subcaption}
\usepackage{multirow}

\usepackage{algorithm}
\usepackage{algorithmic}

\usepackage{newfloat}
\usepackage{listings}
\DeclareCaptionStyle{ruled}{labelfont=normalfont,labelsep=colon,strut=off} 
\floatstyle{ruled}
\newfloat{listing}{tb}{lst}{}
\floatname{listing}{Listing}

\usepackage{booktabs}

\title{FailureSpot: Label-Efficient Timestamp-Level Failure Detection for Vision-Language-Action Models}
\author {
    Jie Ma\textsuperscript{\rm 1}\equalcontrib,
    Zongxi Liu\textsuperscript{\rm 1}\equalcontrib,
    Yi Zhu\textsuperscript{\rm 1}\corresponding
}
\affiliations {
    \textsuperscript{\rm 1}Wayne State University\\
    jie.ma@wayne.edu, ZongxiLiu@wayne.edu, yzhu39@wayne.edu
}

\begin{document}

\maketitle

\begin{abstract}
Vision-language-action (VLA) policies have shown strong potential for general-purpose robotic manipulation, but they can still fail unpredictably during long-horizon execution, making reliable failure detection essential for safe deployment. Existing methods either rely on visual models that typically detect failures only after erroneous actions have occurred, or use lightweight proactive detectors trained on VLA internal representations. However, these proactive methods are often supervised with trajectory-level labels, causing normal pre-failure behavior in unsuccessful trajectories to be incorrectly labeled as failure. This supervision mismatch introduces label noise and limits both trajectory-level detection accuracy and precise timestamp-level failure localization. In this work, we study fine-grained timestamp-level VLA failure detection while addressing the cost of dense annotation. We propose a data-efficient framework that first leverages unlabeled VLA action chunks to construct action-derived weak supervision signals, capturing abnormal patterns such as inconsistent consecutive chunks, frozen or idle actions, and aggressive random motions. We then use active learning to select only the most uncertain trajectories for timestamp-level annotation and fine-tune the detector with these informative labels. Experiments across multiple VLA policies show that our method improves both timestamp-level and trajectory-level failure detection performance.
\end{abstract}


\section{Introduction}
\label{sec:intro}

Vision-language-action (VLA) policies have emerged as a promising paradigm for general-purpose robotic manipulation, enabling robots to follow language instructions and map visual observations directly to action sequences. Despite their strong generalization capabilities, VLA policies can still fail in diverse and unpredictable ways during long-horizon execution~\cite{zhang2026safevla,ying2026agentsafe,xing2026towards,cui2026libero}. For example, a robot may miss the target object, become stuck, remain idle, or produce unstable actions that no longer make meaningful progress toward the task. Accurately detecting such failures is therefore essential for safe and reliable deployment, especially when the robot needs to intervene, recover, or stop execution before causing further errors.

\begin{figure}[t]
\centering
\includegraphics[width=\columnwidth]{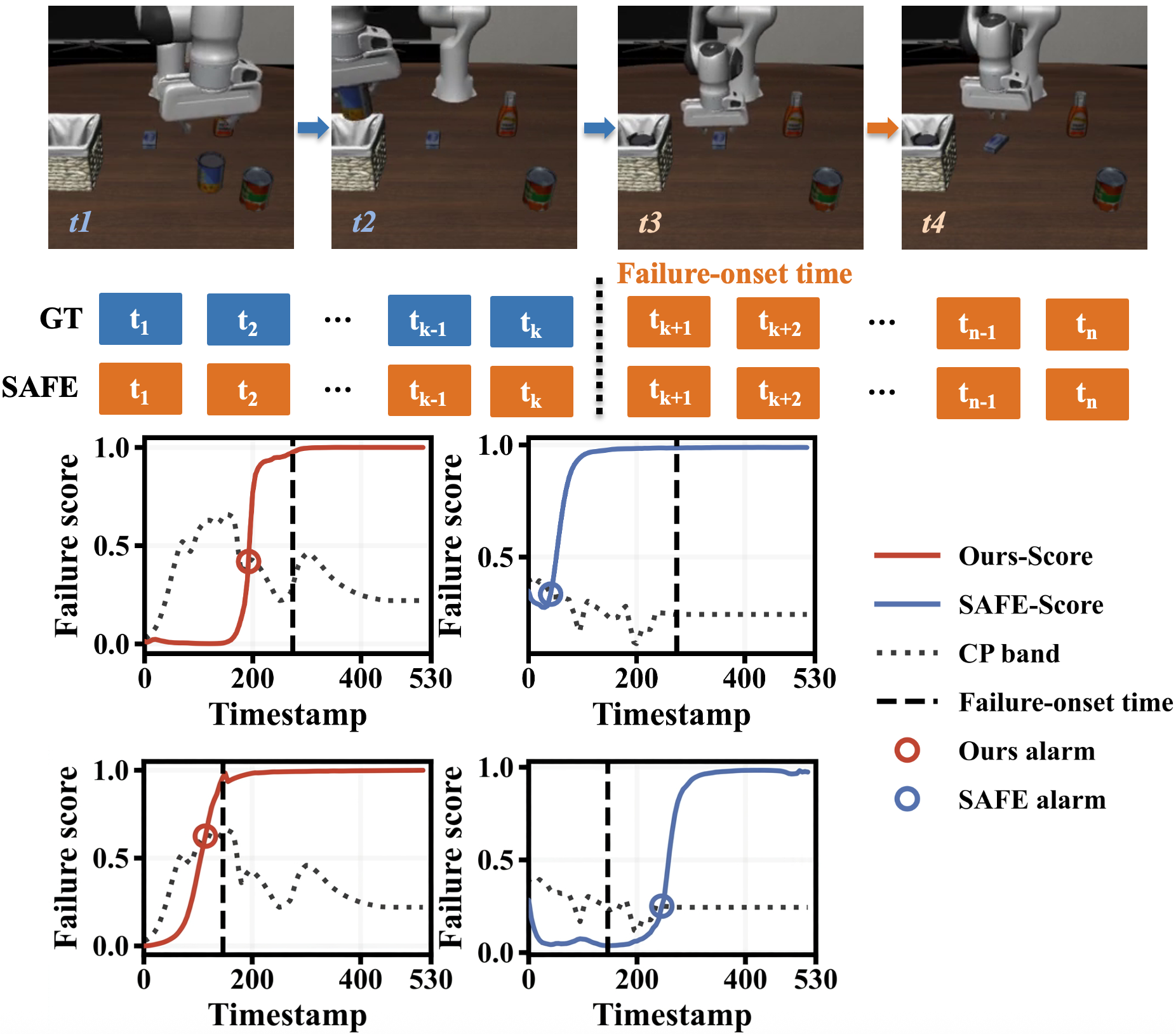}
\caption{\textbf{Timestamp-level failure detection.}
SAFE labels the whole trajectory as failure, which makes the timestamp level detection less accurate: the detected failure-onset time is either too early or too late in the two examples. In contrast, our proposed FailureSpot achieves more accurate timestamp-level failure detection.}
\label{fig:teaser}
\end{figure}

Existing VLA failure detection methods~\citep{du2023vision,duan2024aha} commonly determine whether a failure has occurred by analyzing camera observations of robot execution. Although effective, these approaches are inherently reactive: they can recognize a failure only after the erroneous behavior has already been executed and become visible in the observations. SAFE~\citep{gu2026safe} instead introduces a proactive approach that trains a lightweight detector on the internal representations of a VLA policy. At each timestamp, SAFE takes the current VLA representation as input and outputs a failure score for the corresponding action before that action is executed, making it suitable for runtime monitoring and early intervention.

However, SAFE uses trajectory-level failure labels to supervise timestamp-level detection. As shown in Figure~\ref{fig:teaser}, SAFE treats every timestamp in an unsuccessful trajectory as failure, including the initial timestamps at which the robot is still behaving normally. This supervision mismatch introduces substantial label noise by treating normal pre-failure representations as failure examples. As a result, the detector learns a less discriminative boundary between normal and failure behavior, which can degrade both trajectory-level detection accuracy and timestamp-level failure localization. At the trajectory level, noisy timestamp detections make it more difficult to reliably distinguish successful and failed trajectories; at the timestamp level, they prevent the detector from identifying the precise failure onset. As illustrated by the two examples in Figure~\ref{fig:teaser}, SAFE raises an alarm either too early or too late relative to the ground-truth onset.

In this paper, we address this limitation by studying timestamp-level failure detection for VLA policies. Rather than assigning a single label to an entire trajectory, we annotate robot action chunk at each timestamp with a failure label. Such fine-grained supervision provides a more accurate learning signal for failure detection at each timestamp. However, obtaining timestamp-level labels requires dense manual inspection of robot trajectories, making annotation labor-intensive and costly, particularly for large-scale datasets. The key challenge is therefore to achieve accurate timestamp-level failure detection while minimizing the required human labeling effort.

To address this challenge, we propose FailureSpot, a data-efficient framework that combines action-derived weak supervision with active learning. The core idea is to exploit abundant unlabeled VLA action chunks to obtain approximate timestamp-level supervision and reserve human annotation for only the most informative trajectories. Our framework consists of two stages. First, we construct weak supervision signals from the output action chunks of VLAs, capturing abnormal patterns such as inconsistency between consecutive chunks, frozen or idle actions, and aggressive random motions. These signals are used to pretrain a failure detector without human timestamp-level labels. Second, we estimate the uncertainty of the pretrained detector and select only the most uncertain trajectories for dense annotation. Fine-tuning on such selected labeled subset enables the detector to learn more accurate failure boundaries while substantially reducing annotation cost.

The experimental results on three different VLA policies demonstrate that our method consistently improves both trajectory-level and timestamp-level failure detection over existing baselines. Further ablation studies validate the contributions of both action-derived weak supervision and uncertainty-based active learning, showing that the proposed framework achieves an effective trade-off between failure-onset accuracy and human labeling effort.

Our main contributions are summarized as follows:
\begin{itemize}
\item We study timestamp-level failure detection for VLA and construct fine-grained annotations for training and evaluation.
\item We propose an action-derived weak supervision strategy that leverages consistency and magnitude patterns in VLA action chunks to pretrain a timestamp-level failure detector without requiring dense timestamp-level labels.
\item We use uncertainty-based active learning to select only the most informative trajectories for timestamp-level annotation, substantially reducing labeling cost.
\end{itemize}

\section{Related Work}

\noindent\textbf{Vision-language-action policies.}
VLA policies map a language instruction and visual observations directly to robot
actions~\cite{ma2026survey,zhong2025survey,wang2026unified,intelligence2025pi,li2026pointvla,wen2025tinyvla}. Representative efforts toward scalable and generalist robot policies
include RT-1~\citep{brohan2022rt}, RT-2~\citep{brohan2023rt}, and
Octo~\citep{team2024octo}. They differ partly in how the action is produced.
Autoregressive models
such as OpenVLA~\citep{kim2024openvla} discretize the action space and emit one
action token at a time. Chunked models instead predict a short segment of future
actions in one forward pass and execute only its first few steps before
re-planning, a strategy also used by Action Chunking with Transformers
(ACT)~\citep{zhao2023learning}. $\pi_0$~\citep{black2024pi_0} generates the chunk with flow
matching~\citep{lipman2022flow}, following the same design idea as diffusion
policies~\citep{chi2025diffusion}, while $\pi_0$-FAST~\citep{pertsch2025fast}
compresses the chunk into a small number of tokens so that an autoregressive
backbone can decode it efficiently. These policies are commonly evaluated on
simulated manipulation suites such as LIBERO~\citep{liu2023libero} and
SimplerEnv~\citep{li2024evaluating}. This paper studies failure detection for VLA policies, which aims to determine whether the robot is failing at each
timestamp of an execution.

\noindent\textbf{Failure detection for VLAs.}
Existing VLA failure detection methods can be broadly categorized as reactive or proactive approaches.

\emph{Reactive approaches} rely on observations collected after the robot has
executed its actions. Vision-based methods, for example, feed execution frames from cameras
to a vision-language model to determine whether the task is progressing
normally~\citep{du2023vision,grislain2025failsense,lin2025failsafe} or to explain what has gone
wrong~\citep{duan2024aha}. However, they can generally detect a
failure only after its behavioral consequences have become visible in the
observations, making them unsuitable for early warning. In addition, although these methods benefit from the semantic knowledge of vision-language models, querying a large model at every step is computationally expensive. 

\emph{Proactive approaches} instead operate on information available before the corresponding action is executed. VLMs have been used to predict failures before execution~\citep{yang2026fpc,ma2026cyclevla}, but their high computational cost can limit their use for real-time monitoring. A separate line of work develops lightweight detectors that operate either on the VLA's output action chunks~\cite{agia2024unpacking,huang2026actprobe} or on learned representations~\cite{xu2025can,gu2026safe,zhang2026foresight}. For example, STAC~\citep{agia2024unpacking} samples multiple action chunks and measures disagreement among their overlapping predictions as an indicator of policy uncertainty. ActProbe~\citep{huang2026actprobe} avoids repeated sampling by extracting temporal consistency and action-magnitude signals from action chunks produced by a single policy forward pass and mapping them to timestamp-level failure scores with a lightweight detector. FAIL-Detect~\citep{xu2025can} measures how far a representation lies from the distribution of successful trajectories. SAFE~\citep{gu2026safe} attaches a lightweight MLP or LSTM detector to the internal representations of a VLA and outputs a failure score at every timestamp. Foresight~\citep{zhang2026foresight} detects failures using latent representations from an action-conditioned world model for long-horizon tasks. Because these action outputs or representations are available before execution, such lightweight detectors can provide proactive alarms with relatively low additional inference cost.

However, a key limitation of these methods is the lack of direct timestamp-level supervision and evaluation. They output a failure score at every timestamp but train the detectors by assigning the trajectory-level outcome to all timestamps. Consequently, all timestamps in a failed trajectory are labeled as failures, including the initial portion in which the robot is still behaving normally. This label noise weakens the learned distinction between normal and failure behaviors, degrading both trajectory-level detection accuracy and timestamp-level failure localization. Concurrent work, Hide-and-Seek~\citep{park2026hide}, addresses this supervision mismatch through inter-trajectory and intra-trajectory contrastive learning, which discovers temporally localized failure signals using trajectory-level labels alone. However, its primary detection accuracy is still evaluated at the trajectory level rather than through fine-grained classification of individual timestamps.

Our method also follows the proactive setting and uses VLA internal representations as the detector input. In contrast to prior methods, we explicitly train and evaluate failure detection at the timestamp level. To limit the cost of dense annotation, we use action-derived signals as weak supervision and employ active learning to request timestamp-level annotations only for the trajectories on which the detector is most uncertain. This design directly supervises the failure boundary while substantially reducing human labeling effort.



\section{Method}
\label{sec:method}

\subsection{Problem Setting}
We study timestamp-level failure detection for VLA policies. Let trajectory $i \in {1,\ldots,I}$ contain $T_i$ timestamps. At each timestamp $t \in T_i$, the VLA policy produces an internal representation $h_{i,t}$ and an action chunk to be executed by the robot. Our objective is to determine whether the robot is in a failure state at each timestamp using the policy representations available before action execution. We formulate this problem as timestamp-level binary classification. The detector takes the internal representation $h_{i,t}$ of the VLA policy as input and outputs a failure probability $p_{i,t}=f_{\theta}(h_{i,t}) \in [0,1]$. In this paper, we explore both MLP and LSTM as the lightweight detectors.

\subsection{Timestamp-level failure annotation}
Existing robot datasets typically provide only a success or failure label for the entire trajectory. We therefore introduce timestamp-level annotations for training and evaluation. Each timestamp is assigned a binary label $y_{i,t}\in{0,1}$, where $y_{i,t}=1$ indicates failure and $y_{i,t}=0$ indicates normal execution. A timestamp is labeled as failure when the robot no longer makes meaningful progress toward the task objective. Examples include missing the target object, becoming stuck or idle, moving unpredictably in free space, or producing actions that are inconsistent with the intended task. For a failed trajectory, the failure onset is defined as the first timestamp labeled as failure. This annotation enables training and evaluation for fine-grained failure detection.

\subsection{Framework overview}
Directly training such a detector requires dense timestamp-level annotations in every trajectory, which are expensive to obtain. To reduce this cost, we propose a two-stage label-efficient framework. First, we pretrain the detector using weak supervision derived from the consistency and magnitude of the VLA-predicted action chunks. This stage requires no human timestamp-level annotations. Second, we use the pretrained detector to identify the most uncertain trajectories and request timestamp-level annotations only for this small subset. We then fine-tune the detector with the selected annotations to improve timestamp-level failure detection while limiting human labeling effort.

\subsection{Pretraining with weak supervision}
We observe that many VLA failures are reflected in abnormal patterns in the predicted action chunks. These patterns can appear in different forms. For example, when the robot becomes stuck or idle, the output actions often have small magnitudes and exhibit little temporal variation. When the policy loses coherent control, consecutive action chunks may become inconsistent, and the predicted actions may fluctuate sharply or point in unstable directions. These observations motivate using action-derived signals as weak supervision for timestamp-level failure detection. Specifically, we define three signals: short-term action consistency, full-chunk action magnitude, and executed action magnitude.

Formally, let trajectory $i \in (1,\dots,I)$ contains $T_i$ timestamps, At timestamp $t \in (1,\dots,T_i)$, the VLA outputs an action chunk $\mathbf{A}_{i,t}=[\mathbf{a}^1_{i,t},\dots,\mathbf{A}^H_{i,t}], \mathbf{a}^h_{i,t}\in\mathbb{R}^{D}$, where $\mathbf{a}^h_{i,t}$ denotes the $h$-th action in the chunk, $H$ is the number of actions in each chunk, and $D$ is the action dimension. Each action specifies the robot gripper state, including its position, orientation, and open/close command. Although the VLA predicts $H$ future actions at each timestamp, the robot usually executes the first $K$ actions before receiving a new observation and generating the next action chunk, where $K \leq H$. Autoregressive policies such as OpenVLA~\citep{kim2024openvla} only output and execute one action at a time, so $H=K=1$.

\paragraph{Short-term action consistency.} 
Because consecutive action chunks overlap, their overlapping parts should be consistent~\citep{agia2024unpacking}. We define the short-term action-consistency at timestamp $t$ as the average $\ell_2$ distance between the unexecuted future actions from the previous chunk and the executed actions from the current chunk. Specifically:
\begin{equation}
    c_{i,t} = \frac{1}{M} \sum_{j = 1,\dots,M} || \mathbf{a}_{i,t-1}^{K+j} - \mathbf{a}_{i,t}^j ||_2,
\end{equation}
where $M=\min(K,H-K)$ is the number of overlapping actions being compared. A larger $c_{i,t}$ indicates a stronger inconsistency between consecutive action chunks, suggesting that the VLA policy may be unstable or entering a failure state. For autoregressive policies, in which the consecutive action chunks have not overlap, we define Short-term action consistency as the $\ell_2$ distance between the consecutive actions $c_{i,t} = ||\mathbf{a}_{i,t-1} - \mathbf{a}_{i,t} ||_2$. 

\paragraph{Full-chunk action magnitudes and executed action magnitudes.} Another weak-supervision signal measures the magnitude of the output actions. At timestamp t in trajectory i, each action $\mathbf{a}_{i,t}^j$ is a $D$-dimensional vector. We define the magnitude of an individual action as its $\ell_2$ norm, written as $||\mathbf{a}_{i,t}^j||_2$. We first compute the average magnitude of the whole output actions as: $r_{i,t} = \frac{1}{H} \sum_{j = 1,\dots,H} ||\mathbf{a}_{i,t}^j||_2$. We also compute the average magnitude of the executed output actions as: $r^{exec}_{i,t} = \frac{1}{K} \sum_{j = 1,\dots,K} ||\mathbf{a}_{i,t}^j||_2$. Abnormally small magnitudes may indicate frozen or idle behavior, whereas unusually large or unstable magnitudes may indicate aggressive random actions. We use both signals as the weak supervision. For autoregressive policies, in which the consecutive action chunks have not overlap, the average magnitude of the whole output actions and executed actions are the same, $r_{i,t} = r^{exec}_{i,t} = ||\mathbf{a}_{i,t}||_2$

\paragraph{Exponential moving average.} For each of the three signals, we also compute an exponential moving average to capture the temporal trend of the signal and reduce short-term noise. We define its exponential moving average as:
\begin{equation}
    E(u)_{i,t} = \beta E(u)_{i,t-1} + (1 - \beta) u_{i,t},
\end{equation}
where $u_{i,t}$ represent each signal's value at time $t$ and $E_{i,t}$ is initialized as $E_{i,1}=u_{i,1}$. $\beta$ is the smoothing coefficient. In this paper, we set $\beta = 0.8$.

\paragraph{Combined signal.} We combine the instantaneous value and its exponential moving average of each signal into a single scalar signal value, by taking the weighted sum of them: $s_{i,t}=\big[c_{i,t}+E(c)_{i,t}\big] + \eta_1\big[r_{i,t}+E(r)_{i,t}\big] + \eta_2\big[r^{exec}_{i,t}+E(r^{exec})_{i,t}\big]$,
where $\eta_1$ and $\eta_2$ are hyper-parameters to balance the weights among the three types of signals. 

\paragraph{Normalization.} After obtaining the scalar score $s_{i,t}$, we normalize it using statistics computed only from the training set. Specifically, we compute the $1$st and $99$th percentiles of $s_{i,t}$ over all training timestamps, denoted as $s_{low}$ and $s_{high}$, respectively. We linearly rescale the $s_{i,t}$ scores using $s_{low}$ and $s_{high}$ and clip the results to the range $[0,1]$: 
\begin{equation}
    \tilde{s}_{i,t}=clip(\frac{s_{i,t}-s_{low}}{s_{high}-s_{low}+\epsilon},0,1),
\end{equation}
where $\epsilon$ is used for numerical stability. After normalization, larger $\tilde{s}_{i,t}$ indicates stronger evidence of failure, which serves as the final soft timestamp-level pseudo-label for training the MLP and LSTM.

\subsection{Fine-tuning with active learning}
Weak supervision provides a scalable initialization, but action-consistency pseudo-labels can be noisy. We therefore use active learning~\cite{settles2009active} to acquire a small set of informative timestamp-level annotations for fine-tuning. Starting from the weakly pretrained detector, we estimate the uncertainty of each timestamp using the entropy of the detected failure probability: $H(p_{i,t})=-p_{i,t}\log p_{i,t}-(1-p_{i,t})\log(1-p_{i,t})$, where $p_{i,t}$ is the detector-detected failure probability for timestamp $t$ in trajectory $i$. Because timestamp-level annotation is more practical when performed over complete trajectories, we select trajectories rather than isolated timestamps. 

The uncertainty score of a trajectory is defined as the average entropy across all timestamps: $U_i=\frac{1}{T_i}\sum_{t=1}^{T_i}H(p_{i,t})$, where $T_i$ is the length of trajectory $i$. We then rank all unlabeled trajectories by their uncertainty scores and request timestamp-level annotations only for the top-ranked trajectories. This avoids labeling the full dataset and focuses human effort on the trajectories where the detector is most uncertain and where annotation is expected to be most informative.

After annotation, we fine-tune the detector using the selected timestamp-level labels with a supervised binary classification loss. This two-stage design combines the scalability of action-derived weak supervision with the accuracy of targeted human annotation, allowing the detector to learn fine-grained failure boundaries while requiring substantially fewer timestamp-level labels.

\subsection{Conformal prediction}
The proposed failure detector outputs a failure score at each timestamp. To convert these scores into binary failure alarms, we trigger an alarm whenever the predicted score exceeds a decision threshold. Following SAFE~\citep{gu2026safe}, we determine this threshold using functional conformal prediction~\citep{diquigiovanni2021importance}. We construct a one-sided, time-varying conformal
band $b_{\alpha}(t)$ using held-out successful trajectories from the seen
tasks, so that the threshold adapts to the natural temporal variation of
failure scores under normal execution. For a significance level $\alpha$, a timestamp $t$ in trajectory $i$ is classified as failure if the predicted score satisfies $p_{i,t}\geq b_{\alpha}(t)$. The failure-onset time is
defined as
\begin{equation}
    \tau_{\alpha}=\min\{t:p_{i,t}\geq b_{\alpha}(t)\}.
\end{equation}

\section{Evaluation}
\label{sec:experiment}

\subsection{Experimental settings}

\noindent\textbf{Benchmarks.}
We evaluate on LIBERO-10 (LIBERO-Long)~\citep{liu2023libero}, a suite that contains 10 long-horizon
manipulation tasks. We consider three representative VLA policies: $\pi_0$~\citep{black2024pi_0},
$\pi_0$-FAST~\citep{pertsch2025fast}, and OpenVLA~\citep{kim2024openvla}. At each step, $\pi_0$ predicts an action chunk of length
$H=50$ and $\pi_0$-FAST predicts an action chunk of length $H=10$; both execute the
first $K=5$ actions before re-planning. OpenVLA is an autoregressive policy that
outputs and executes a single action per step ($H=K=1$). The three policies share the
same action space (7 dimensions, corresponding to end-effector position, orientation,
and gripper). For each policy, we roll out 50 trajectories on each of the 10 tasks,
yielding 500 trajectories in total; the final success rates of the three policies are
$81.4\%$, $61.4\%$, and $53.8\%$, respectively. All experiments are run on two NVIDIA RTX A6000 GPUs with 48GB memory.

\noindent\textbf{Timestamp-level annotation.}
Since existing datasets do not provide timestamp-level failure annotations, we manually
inspect each trajectory frame by frame and annotate the timestamp at which the robot
first clearly deviates from the task objective (e.g., missing the target object, staying
idle, getting stuck, or moving randomly in free space), denoted as the failure-onset time
$t_{\mathrm{dev}}$. If the robot returns to normal execution after the deviation and
eventually completes the task, we additionally annotate its recovery time
$t_{\mathrm{rec}}$. Accordingly, each trajectory $i$ has both a trajectory-level label $Y_i$
($Y_i=1$ indicates final failure and $Y_i=0$ indicates final success) and per-timestamp labels $y_{i,t}$:
\begin{itemize}
    \item For final-failure trajectories ($Y_i=1$), all timestamps with $t\ge t_{\mathrm{dev}}$ are labeled as 1;
    \item For trajectories that recover to success after deviation ($Y_i=0$)m only timestamps in
          $t_{\mathrm{dev}}\le t<t_{\mathrm{rec}}$ are labeled as 1;
    \item For trajectories that are normal throughout ($Y_i=0$): all timestamps are labeled 0.
\end{itemize}
The timestamp-level annotation enables the identification of recovered-success trajectories where the final trajectory is success $Y_i=0$ while some timestamps still
have $y_{i,t}=1$.

\noindent\textbf{Data splits.}
We use three tasks as seen tasks and reserve the remaining seven as
unseen evaluation tasks. Among the 50 trajectories of each seen task, $60\%$ are used for training and the remaining $40\%$ serve as the seen-task evaluation set; all 350 trajectories of the 7 unseen tasks are used only for evaluation. The budget for timestamp-level annotation is fixed at 15\% of the trajectories from the seen tasks, with all selected trajectories drawn from the training set. All results are averaged over the three seeds.


\noindent\textbf{Baselines.}
We compare against three groups of baselines:

\noindent\textit{Training-free detectors.}
\textbf{STAC-Single}~\citep{agia2024unpacking} is a real-time, single-sample
variant of STAC that measures disagreement between the overlapping regions of
consecutive action chunks without requiring training or labeled data. It is
not applicable to OpenVLA because OpenVLA outputs only one non-overlapping
action at each timestamp.

\noindent\textit{Detectors with trajectory-level labels.}
These methods use only trajectory-level labels. \textbf{LogpZO}, introduced as part of
\textbf{FAIL-Detect}~\citep{xu2025can}, compares the representation
distributions of successful and failed trajectories to produce timestamp-level
failure scores. \textbf{SAFE-MLP} and
\textbf{SAFE-LSTM}~\citep{gu2026safe} instead learn timestamp-level failure
scores directly from VLA representations using trajectory-level labels.
\textbf{ActProbe}~\citep{huang2026actprobe} maps
consistency and magnitude statistics of a single predicted action chunk to a
timestamp-level score with a lightweight detector. We only evaluate its performance on $\pi_0$ and OpenVLA due to open-sourced code availability. 

\noindent\textit{Detectors with timestamp-level labels.}
We directly train an \textbf{MLP} and \textbf{LSTM}~\citep{hochreiter1997long} on the same VLA representations using timestamp-level annotations, without weak-supervision pretraining or active selection. For a fair comparison, both baselines use the same timestamp-level annotation budget as our method (15\% of the training trajectories).

\noindent\textbf{Evaluation Metrics.}
We report both timestamp-level and trajectory-level AUROC, which measure the ranking quality of the detector across failure and non-failure samples without requiring a fixed decision threshold.


To evaluate the accuracy given the decision threshold, we sweep 15 significance levels $\alpha\in[0.02,0.9]$ and evaluate the
timestamp-level balanced accuracy:
\begin{equation}
    \mathrm{BalACC}_{\mathrm{ts}}(\alpha)
    =
    \frac{1}{2}
    \left(
        \mathrm{TPR}_{\mathrm{ts}}(\alpha)
        +
        \mathrm{TNR}_{\mathrm{ts}}(\alpha)
    \right).
\end{equation}

\subsection{Overall Performance}
\begin{table*}[h]
\centering
\small
\caption{Overall performance. The best result in each column is highlighted in
bold, and the second-best result is underlined.}
\begin{tabular}{lllllllllllll}
\toprule
& \multicolumn{4}{c}{$\pi_0$}
& \multicolumn{4}{c}{$\pi_0$-FAST}
& \multicolumn{4}{c}{OpenVLA} \\
\cmidrule(lr){2-5}
\cmidrule(lr){6-9}
\cmidrule(lr){10-13}

& \multicolumn{2}{l}{Trajectory-level}
& \multicolumn{2}{l}{Timestamp-level}
& \multicolumn{2}{l}{Trajectory-level}
& \multicolumn{2}{l}{Timestamp-level}
& \multicolumn{2}{l}{Trajectory-level}
& \multicolumn{2}{l}{Timestamp-level} \\
\cmidrule(lr){2-3}
\cmidrule(lr){4-5}
\cmidrule(lr){6-7}
\cmidrule(lr){8-9}
\cmidrule(lr){10-11}
\cmidrule(lr){12-13}

Methods
& Seen & Unseen & Seen & Unseen
& Seen & Unseen & Seen & Unseen
& Seen & Unseen & Seen & Unseen \\
\midrule

SAFE-MLP
& 78.4 & 67.7 & 82.8 & 76.6
& 68.3 & 69.5 & 77.5 & 81.9
& 60.4 & 54.5 & 61.3 & 64.7 \\

SAFE-LSTM
& 59.5 & 63.2 & 68.7 & 70.0
& 83.1 & 73.4 & 65.6 & 65.2
& \textbf{61.9} & \underline{57.2} & 64.1 & \textbf{67.7} \\

STAC
& 47.7 & 64.2 & 65.9 & 64.2
& 84.4 & 73.3 & 75.8 & 83.9
& -- & -- & -- & -- \\

LogpZO
& 75.0 & 67.3 & 78.8 & 69.8
& \underline{89.8} & \underline{81.1} & 68.6 & 74.5
& 55.7 & 54.8 & 64.4 & 63.4 \\

ActProbe
& 78.8 & 69.5 & 66.7 & 55.4
& -- & -- & -- & --
& 56.6 & \textbf{58.5} & 67.1 & 64.6 \\

MLP
& 62.8 & 58.6 & 79.9 & 61.8
& 70.6 & 79.5 & 78.0 & 78.6
& 40.4 & 41.1 & 42.6 & 38.0 \\

LSTM
& 61.5 & 63.9 & 78.3 & 70.7
& 68.3 & 69.0 & 71.8 & 73.1
& 47.1 & 47.9 & 50.9 & 45.0 \\

\textbf{FailureSpot-MLP}
& \textbf{86.2} & \textbf{81.8}
& \textbf{95.3} & \textbf{90.5}
& \textbf{90.8} & \textbf{82.6}
& \textbf{81.2} & \textbf{85.8}
& 50.6 & 50.3
& \underline{74.5} & 63.0 \\

\textbf{FailureSpot-LSTM}
& \underline{82.6} & \underline{78.6}
& \underline{92.1} & \underline{84.3}
& 76.3 & 78.3
& \underline{78.5} & \underline{84.4}
& \underline{61.4} & 56.1
& \textbf{75.0} & \underline{66.5} \\

\bottomrule
\end{tabular}
\label{tab:overall}
\end{table*}

Table~\ref{tab:overall} summarizes the trajectory-level and timestamp-level
AUROC of our method and the baselines. With only a 15\% timestamp-level annotation budget, our method achieves the highest overall timestamp-level AUROC across all three VLA policies, demonstrating that the proposed training framework can learn fine-grained failure signals with limited human annotation.

Our method also achieves the best trajectory-level AUROC on $\pi_0$ and $\pi_0$-FAST, while performing slightly worse on OpenVLA. This may be because OpenVLA outputs only a single action at each timestamp, consecutive predictions contain no overlapping actions, and the short-term action-consistency signal cannot be fully exploited.

\begin{figure}[t]
\centering
\includegraphics[width=1.0\columnwidth]{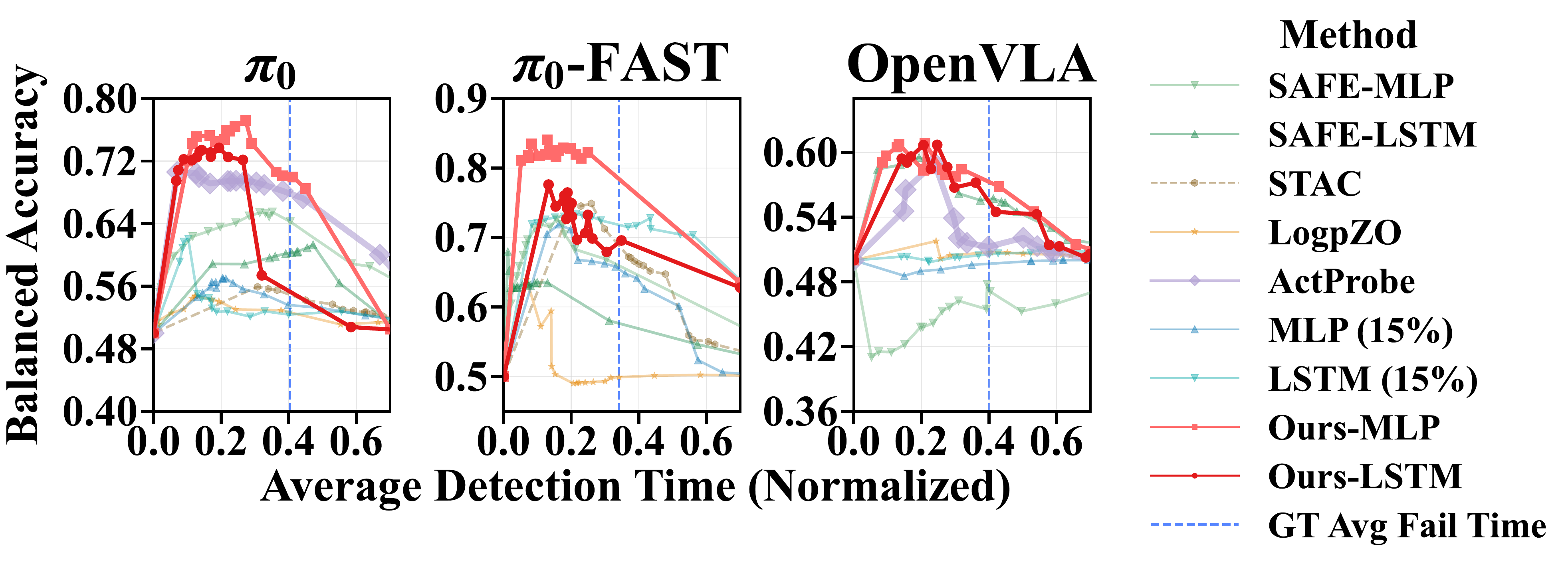}
\caption{Balanced accuracy and detected failure-onset time on unseen tasks. The y-axis is
timestamp-level balanced accuracy and the x-axis the average normalized
detected failure-onset time on failed trajectories; each marker is one conformal level
$\alpha$. The blue dashed line marks the mean ground-truth failure-onset time.}
\label{fig:tradeoff}
\end{figure}

\subsection{Accuracy and Failure-Onset Time}

Figure~\ref{fig:tradeoff} evaluates the balanced accuracy and detected failure-onset time together under difference significant level $\alpha$ of conformal prediction (different decision thresholds). A failure detector with optimal value of $\alpha$ should achieve high detection accuracy while identifying the failure onset as close as possible to, or slightly before, the ground-truth failure-onset time.


Under this criterion, our method achieves the best overall performance. As shown in Figure~\ref{fig:tradeoff}, it maintains high balanced accuracy across a range of $\alpha$ values. At the $\alpha$ that yields the highest balanced accuracy, our detector triggers failure alarms before the average ground-truth failure-onset time (blue dashed line), enabling timely intervention without sacrificing detection accuracy.

\subsection{Ablation Study}

We isolate the contributions of action-derived weak supervision and active learning. Specifically, we compare: (1) pretraining only with action-derived weak-supervision, (2) pretraining with action-derived weak-supervision followed by fine-tuning with randomly selected timestamp-level data, (3) pretraining with trajectory-level labels followed by active-learning-based fine-tuning, and (4) our full framework combining action-derived weak-supervision pretraining with active-learning-based fine-tuning, using both MLP and LSTM detectors. For all methods, pretraining and fine-tuning use 60\% and 15\% of training trajectories.

As shown in Table~\ref{tab:ablation}, action-derived pretraining alone is insufficient, while adding timestamp-level supervision substantially improves performance. Active learning further improves the effectiveness of the limited annotation budget. Overall, our full method achieves the best timestamp-level AUROC across all three VLA policies and the strongest trajectory-level performance on $\pi_0$ and $\pi_0$-FAST, demonstrating the complementary benefits of action-derived weak supervision and active learning.

\begin{table*}[h]
\centering
\caption{Ablation study. The best result in each column is highlighted
in bold, and the second-best result is underlined.}
\small
\begin{tabular}{llrrrrrr}
\toprule
& & \multicolumn{3}{c}{Trajectory-level}
& \multicolumn{3}{c}{Timestamp-level} \\
\cmidrule(lr){3-5}
\cmidrule(lr){6-8}
Model & Methods & Seen & Unseen & Avg. & Seen & Unseen & Avg. \\
\midrule

\multirow{8}{*}{$\pi_0$}
& action-derived weak-supervision only (MLP)
& 42.1 & 30.4 & 36.3 & 36.7 & 30.7 & 33.7 \\

& action-derived weak-supervision only (LSTM)
& 36.5 & 35.4 & 36.0 & 42.4 & 42.0 & 42.2 \\

& action-derived weak-supervision + random fine-tuning (MLP)
& 79.4 & 75.1 & 77.3 & 80.6 & 78.3 & 79.5 \\

& action-derived weak-supervision + random fine-tuning (LSTM)
& 71.6 & 64.8 & 68.2 & 80.2 & 70.9 & 75.6 \\

& trajectory-level labels + active-learning fine-tuning (MLP)
& \textbf{86.8} & 73.0 & 79.9
& \underline{93.8} & 80.1 & 87.0 \\

& trajectory-level labels + active-learning fine-tuning (LSTM)
& 82.2 & 73.6 & 77.9 & 78.2 & 75.2 & 76.7 \\

& \textbf{FailureSpot-MLP}
& \underline{86.2} & \textbf{81.8} & \textbf{84.0}
& \textbf{95.3} & \textbf{90.5} & \textbf{92.9} \\

& \textbf{FailureSpot-LSTM}
& 82.6 & \underline{78.6} & \underline{80.6}
& 92.1 & \underline{84.3} & \underline{88.2} \\

\midrule

\multirow{8}{*}{$\pi_0$-FAST}
& action-derived weak-supervision only (MLP)
& 78.5 & 73.7 & 76.1 & 68.8 & 73.3 & 71.1 \\

& action-derived weak-supervision only (LSTM)
& 79.2 & 73.8 & 76.5 & 71.7 & 76.0 & 73.9 \\

& action-derived weak-supervision + random fine-tuning (MLP)
& 83.1 & 71.7 & 77.4 & 71.3 & 73.3 & 72.3 \\

& action-derived weak-supervision + random fine-tuning (LSTM)
& 78.9 & 68.5 & 73.7 & 71.5 & 75.5 & 73.5 \\

& trajectory-level labels + active-learning fine-tuning (MLP)
& \underline{88.1} & \textbf{83.0} & \underline{85.6}
& \underline{80.1} & 80.3 & 80.2 \\

& trajectory-level labels + active-learning fine-tuning (LSTM)
& 78.5 & 76.2 & 77.4 & 77.0 & 79.7 & 78.4 \\

& \textbf{FailureSpot-MLP}
& \textbf{90.8} & \underline{82.6} & \textbf{86.7}
& \textbf{81.2} & \textbf{85.8} & \textbf{83.5} \\

& \textbf{FailureSpot-LSTM}
& 76.3 & 78.3 & 77.3
& 78.5 & \underline{84.4} & \underline{81.5} \\

\midrule

\multirow{8}{*}{OpenVLA}
& action-derived weak-supervision only (MLP)
& 32.9 & 38.2 & 35.6 & 52.8 & 61.3 & 57.1 \\

& action-derived weak-supervision only (LSTM)
& 38.1 & 47.4 & 42.8 & 51.0 & 64.8 & 57.9 \\

& action-derived weak-supervision + random fine-tuning (MLP)
& 43.2 & 47.9 & 45.6 & 57.1 & 63.9 & 60.5 \\

& action-derived weak-supervision + random fine-tuning (LSTM)
& 42.4 & 49.0 & 45.7 & 58.7 & 64.2 & 61.5 \\

& trajectory-level labels + active-learning fine-tuning (MLP)
& \textbf{62.9} & 55.5 & \underline{59.2}
& 59.3 & 58.2 & 58.8 \\

& trajectory-level labels + active-learning fine-tuning (LSTM)
& 59.1 & \textbf{64.4} & \textbf{61.8}
& 68.6 & \underline{66.2} & 67.4 \\

& \textbf{FailureSpot-MLP}
& 50.6 & 50.3 & 50.5
& \underline{74.5} & 63.0 & \underline{68.8} \\

& \textbf{FailureSpot-LSTM}
& \underline{61.4} & \underline{56.1} & 58.8
& \textbf{75.0} & \textbf{66.5} & \textbf{70.8} \\

\bottomrule
\end{tabular}
\label{tab:ablation}
\end{table*}

\subsection{Impact of training data amount}
\begin{figure}[t]
\centering
\includegraphics[width=\columnwidth]{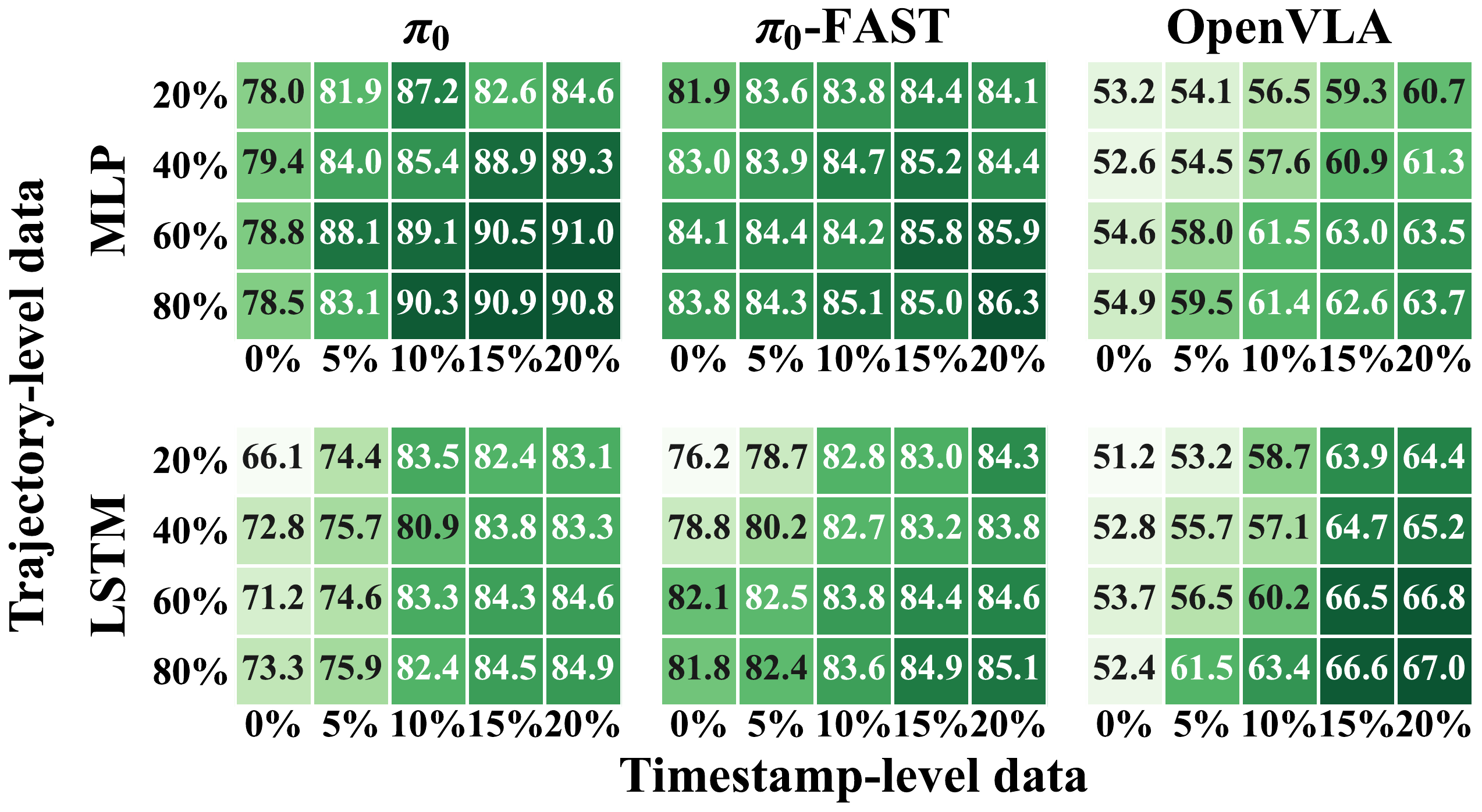}
\caption{Timestamp-level AUROC on unseen tasks w.r.t. training data amount. Rows vary the trajectory-level data
used for weak-supervision pretraining and columns the timestamp-level annotation
used for active fine-tuning; darker green indicates higher AUROC.}
\label{fig:data_scaling}
\end{figure}

We next study how performance scales with the amount of weak-supervision pretraining data and timestamp-level fine-tuning data. As shown in Figure~\ref{fig:data_scaling}, we jointly vary the fraction of trajectories used for pretraining from 20\% to 80\% and the timestamp-level annotation budget from 0\% to 20\%, training a separate model for each configuration. Increasing either source of data improves performance, but the gains from timestamp-level annotations are substantially larger. The configuration used in our main experiments: 60\% of trajectories for pretraining and 15\% timestamp-level annotated trajectories, provides a favorable trade-off between labeling cost and detection performance.

\subsection{Visualization of timestamp-level failure detection}
\begin{figure}[t]
\centering
\includegraphics[width=\columnwidth]{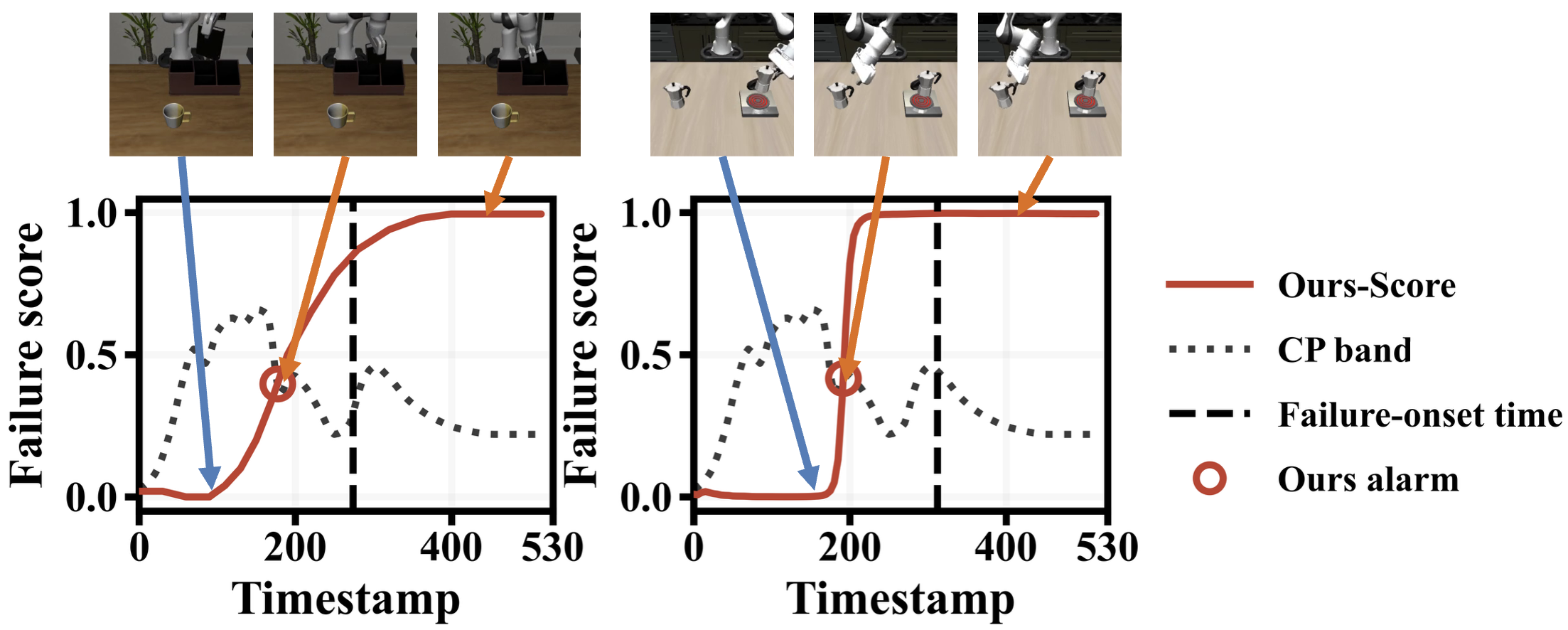}
\caption{\textbf{Examples of timestamp-level failure detection.} Black dotted curves are the threshold calculated by conformal prediction, red circles show the detected failure onset, and black dashed lines the annotated
failure-onset times.}
\label{fig:first_failure}
\end{figure}

Figure~\ref{fig:first_failure} presents two qualitative examples for $\pi_0$. In both trajectories, the detected failure score remains low during normal robot actions, then rises sharply above the decision threshold. The red circles show the detected failure-onset time, which are shortly before the ground-truth failure-onset time (dashed lines). The corresponding camera frames illustrate this transition: frames before the failure onset show continued progress toward the task objective, whereas frames after the failure onset exhibit the behavior that ultimately leads to failure. This indicates that the detector can distinguish the normal robot actions from the emerging failures early enough to support timely intervention.

\begin{figure}[t]
\centering
\includegraphics[width=\columnwidth]{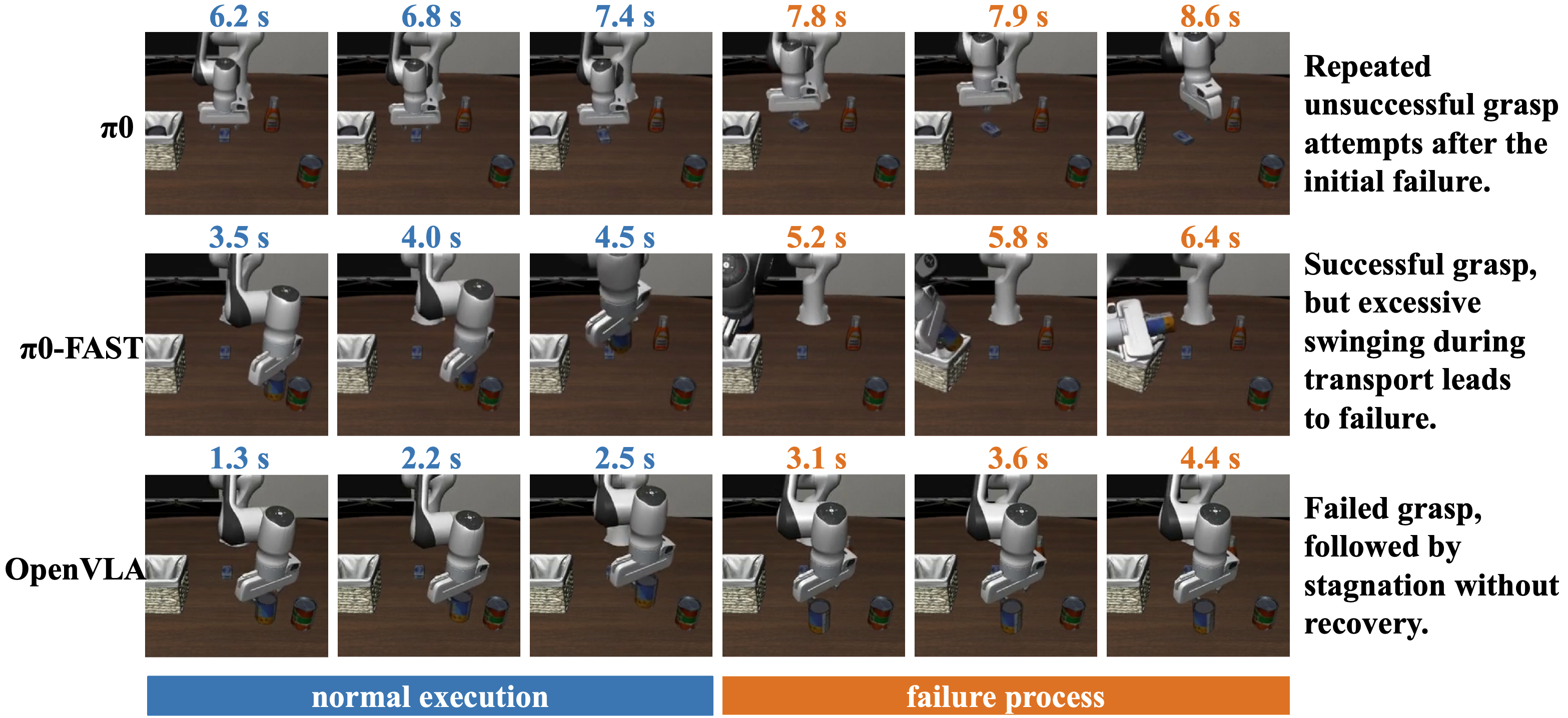}
\caption{Representative failure modes for $\pi_0$, $\pi_0$-FAST, and OpenVLA. Blue timestamps mark the normal execution phase and orange the failure phase.}
\label{fig:failure_modes}
\end{figure}

\subsection{Failure mode discovery}
Inspecting the behaviors of failed trajectories reveals three recurring failure modes. The first is repeated ineffective actions, where the robot repeatedly executes similar action patterns, such as repeated grasp attempts, without making progress toward the task. In this case, consecutive action chunks exhibit strong temporal similarity despite persistent task failure. The second is stagnation, where the gripper settles into a nearly fixed pose and stops making meaningful progress; correspondingly, the generated action magnitudes become close to zero. The third is ineffective large-amplitude motion, where the robot continues moving substantially without advancing the task; in this case, the action magnitudes remain large while consecutive predictions become less consistent. Together, these failure modes motivate the use of both action magnitude and temporal consistency as complementary weak-supervision signals.

We also observe that different VLA policies tend to exhibit different failure modes. As shown in Figure~\ref{fig:failure_modes}, $\pi_0$ commonly exhibits repeated but unsuccessful grasp attempts, $\pi_0$-FAST more often produces excessive swinging motions, and OpenVLA frequently enters stagnation.

\section{Conclusion}

We study timestamp-level failure detection for VLA policies, with the goal of identifying not only whether a trajectory fails but also when the failure begins. We propose FailureSpot, a label-efficient framework that combines action-derived weak supervision with active learning to reduce the cost of timestamp-level annotation. Experiments across multiple VLA policies show improved timestamp-level and trajectory-level detection performance, together with more accurate and timely failure alarms. Future work will explore how the proposed timestamp-level failure detector can be used for timely failure intervention and recovery.

\bibliography{aaai2027}


\end{document}